\documentclass[11pt]{article}

\usepackage{jmlr2e}
\usepackage{booktabs}
\usepackage{enumitem}
\usepackage[usenames,dvipsnames,x11names]{xcolor}
\usepackage{booktabs}
\usepackage{mathtools}
\usepackage{bbm}

\usepackage{natbib,array,float,makecell,pdflscape,subcaption}
\allowdisplaybreaks
\usepackage{color,charter,graphicx,natbib,amsmath,amssymb,amsfonts,setspace,multirow,caption}

\usepackage{setspace}

\ShortHeadings{Robust DNN Surrogate Models with Uncertainty Quantification}{Zhang and Li}
\firstpageno{1}
\begin{document}



\title{Robust DNN Surrogate Models with Uncertainty Quantification via Adversarial Training}
\author{\name Lixiang Zhang \email lzz46@psu.edu
       \AND
       \name Jia Li \email jiali@psu.edu \\
       \addr Department of Statistics\\
       The Pennsylvania State University\\
       University Park, PA 16802, USA}

\editor{}

\maketitle

\begin{abstract}
For computational efficiency, surrogate models have been used to emulate mathematical simulators for physical or biological processes. High-speed simulation is crucial for conducting uncertainty quantification (UQ) when the simulation is repeated over many randomly sampled input points (aka, the Monte Carlo method). In some cases, UQ is only feasible with a surrogate model. Recently, Deep Neural Network (DNN) surrogate models have gained popularity for their hard-to-match emulation accuracy. However, it is well-known that DNN is prone to errors when input data are perturbed in particular ways, the very motivation for adversarial training. In the usage scenario of surrogate models, the concern is less of a deliberate attack but more of the high sensitivity of the DNN's accuracy to input directions, an issue largely ignored by researchers using emulation models. In this paper, we show the severity of this issue through empirical studies and hypothesis testing. Furthermore, we adopt methods in adversarial training to enhance the robustness of DNN surrogate models. Experiments demonstrate that our approaches significantly improve the robustness of the surrogate models without compromising emulation accuracy. 
\end{abstract}


\section{Introduction}
\label{intro}

In science and engineering, many computational models with high complexity have been used to simulate physical or biological processes, perform forecasting, and help decision-making.
Examples include global climate models developed in earth system science to predict future climate~\citep{sunyer2014bayesian,qian2016uncertainty}, 
infectious disease models, e.g., stochastic HIV simulator for studying the effects of partnership concurrency on HIV transmission~\citep{andrianakis2015bayesian}, and stochastic and high-dimensional simulators of physics systems based on stochastic partial differential equation (SPDE)~\citep{tripathy2018deep,luo2019deep}. The accuracy of a simulator is measured by how well it reproduces empirical data, and the simulation result often comes with an assessment of uncertainty. Frequently in practice, the input to the simulator contains many parameters, which are not given precisely. The randomness in input causes variation in the output. It is thus desirable to quantify the amount of output uncertainty, e.g., by providing a prediction interval (PI), and even more comprehensively, to specify a probability density function (PDF) for the output.

The most straightforward approach to quantifying uncertainty is to perform Monte Carlo (MC) simulations~\citep{mooney1997monte}. Specifically, random samples of the input parameters are drawn from the vicinity of a given configuration, and simulation is repeated at each input sample. The collection of simulation results is used for uncertainty analysis, e.g., estimating the distribution of the output. The sample size needed for MC to converge can be large, while each round of simulation is often computationally costly. Thus the total amount of computation for the MC simulation can become infeasible. To overcome this hurdle, researchers often adopt a surrogate model, which emulates the original simulator but generates results much faster. In the literature, several different types of surrogate models have been proposed, including Gaussian Process (GP)~\citep{bilionis2012multi,bilionis2013multi}, Polynomial Chaos Expansion (PCE)~\citep{xiu2002modeling,xiu2003modeling,najm2009uncertainty}, and Deep Neural Network (DNN)~\citep{zhu2018bayesian,kabir2018neural,tripathy2018deep,luo2019deep}. Among those models, DNN is attracting growing attention for several reasons. First, DNN can effectively handle high dimensional input frequently encountered in modern applications. Secondly, DNN often yields state-of-the-art accuracy in terms of approximating the simulators. At last, with the rapid advance of computer hardware and optimization techniques such as Adam~\citep{kingma2014adam}, the time to train a DNN has been reduced significantly. 

It is known in image analysis and computer vision that perturbation on the data along some directions can quickly flip the classification by a DNN, even when the human eye cannot notice any change~\citep{goodfellow2014explaining}. This issue has inspired research on the topic of attacks and defenses for DNN~\citep{yuan2019adversarial,zhang2019adversarial}, for example, to mention just a few, Generative Adversarial Network (GAN)~\citep{goodfellow2014generative,creswell2018generative,creswell2018inverting} and Adversarial Autoencoder (AAE)~\citep{makhzani2015adversarial,creswell2018denoising,ge2019dual}. 

Does the high sensitivity of DNN to small input changes in some directions have a severe impact on the uncertainty quantification of the predictions of surrogate models? After all, inputs to surrogate models are not images. In this paper, we answer this question by comparing density functions and hypothesis testing. We also explore whether typical MC simulations are adequate for detecting directional sensitivity. Our study confirms the concern that the emulation error of DNN can increase sharply at a slight change in input depending on the direction of the change, and to further complicate the issue, such errors are not easy to unveil through standard MC sampling. These findings call the attention of researchers using surrogate models to an overlooked yet important issue and shed light on the best practice for emulating simulators.
Furthermore, we propose a computationally efficient adversarial training method and demonstrate that a DNN trained by this method achieves greatly improved consistency in performance without loss in average accuracy. The new method employs a revised objective function applicable to any DNN architecture. Hence, it can serve as a general framework for developing robust surrogate models.

We refer to the survey~\citep{ren2020adversarial} for a thorough review on adversarial learning.
Based on the information available to the adversary, adversarial attacks fall into three types: white-box, gray-box, and black-box attacks~\citep{ren2020adversarial}. In white-box attacks, the adversary has full access to the target model, {\it e.g.}, model architecture, the training algorithm, and the gradients of the prediction function. Such attacks often achieve remarkable degradation in performance~\citep{carlini2017towards}. In gray-box attacks, an extra training process is evoked, during which the target model is accessible. After that training step, the adversary generates adversarial examples without querying the target model~\citep{xu2021grey}. In the most challenging case of black-box attacks, without any knowledge about the target model, the adversary only uses information about the settings or past inputs to identify the vulnerability~\citep{chakraborty2018adversarial}. On the other hand, we can also categorize adversarial attacks into targeted versus non-targeted attacks~\citep{ren2020adversarial}. Targeted attacks attempt to mislead the model to a particular wrong label or direction, while non-targeted attacks aim at causing incorrect prediction. 


The rest of the paper is organized as follows. In Section~\ref{solution}, we present an approach to improve the robustness of the DNN surrogate model using adversarial learning. In Section~\ref{exp}, we describe 
the study on the sensitivity of the DNN surrogate model to perturbation directions and provide experimental results on the performance of our proposed method in terms of both prediction accuracy and uncertainty quantification. Finally, we conclude in Section~\ref{conclude}.

\section{Enhance DNN Surrogate Models by Adversarial Training}
\label{solution}
Consider a computational model, referred to as a simulator, that characterizes a natural process. We generally view the model as a function $F: \mathcal{X} \rightarrow \mathcal{Y}$, where the input $x \in \mathcal{X}$ can contain a great variety of quantities, {\it e.g.}, material properties,  boundary conditions, and initial conditions. The function $F$ is usually so complex that it requires numerical solutions to sophisticated mathematical systems, {\it e.g.}, PDEs. We call the calculation of $F(x)$ querying the simulator. Given the distribution of input $x$, in uncertainty quantification (UQ), we aim at obtaining some statistics about the output $F(x)$, such as mean $E[F(x)]$ and variance $Var[F(x)]$, or more comprehensive description such as the PDF of $F(x)$. As discussed previously, it is computationally intensive or even infeasible to conduct Monte Carlo (MC) simulations via the original simulator. A fast surrogate model, denoted by $\hat{F}$, can be used to approximate $F$. The performance of the surrogate model is measured by the closeness of its solution to the simulator's, which we call {\it emulation accuracy}. We will study the performance of DNN as a surrogate model, in particular, the directional sensitivity of the emulation accuracy.

\subsection{Identify Adversarial Directions}
Since our purpose is to build a robust DNN surrogate model that can avoid drastic accuracy loss with small perturbation in the input, in the terminology of adversarial learning, the ``attacks'' belong to the white-box type, that is, the model is always available when generating a perturbed input. We face the case of non-targeted white-box attacks in a figurative sense. Let $\hat{F}$ be the trained DNN prediction function and $x$ be the input. Finding the most vulnerable perturbation that changes the prediction is equivalent to generating an adversarial example $x^{\prime}=x+\delta$ by adding a perturbation $\delta$ optimized by the following problem:
\begin{equation}
\underset{\delta}{\arg \min }\{\|\delta\|: \hat{F}(x+\delta) \neq \hat{F}(x)\}.
\end{equation}

There are several efficient approaches to the above optimization problem, the most popular being {\it Fast Gradient Sign Method} ({\it FGSM})~\citep{goodfellow2014explaining}. FGSM calculates the gradient of the loss function $J$ with respect to the DNN and creates adversarial examples by the following equation:
\begin{equation}
x^{\prime}=x+\epsilon \cdot \operatorname{sign}\left[\nabla_{x} J(\theta, x, y)\right],
\label{fgsm}
\end{equation}
where $\epsilon$ is the magnitude of the perturbation, $\theta$ contains model parameters, and $y$ is the true output. FGSM has several variants, {\it e.g.}, Target Class Method, and Basic Iterative Method~\citep{kurakin2016adversarial}. However, Cheng {\it et al.} \cite{cheng2021fast} argues that the sign alone may not produce perturbations most effectively. The reason is that the sign only specifies gradients in terms of $\{0,1,-1\}$ and a perturbation generated based on the sign can differ considerably from the gradient (the fastest-changing direction). Hence they proposed the {\it Fast Gradient Non-sign Method} ({\it FGNM}):
\begin{equation}
x^{\prime}=x+\epsilon \cdot \frac{\|\operatorname{sign}\left[\nabla_{x} J(\theta, x, y)\right] \|}{\| \nabla_{x} J(\theta, x, y) \|}  \cdot \nabla_{x} J(\theta, x, y),
\label{fgnm}
\end{equation}
where the perturbation has the same norm as FGSM but follows exactly the direction of the gradient. In our experiments, we use both FGSM and FGNM to generate adversarial samples for the DNN surrogate model and evaluate accuracy on these samples. For commonly used DNN surrogate models, we find that both methods generate adversarial samples with a strong effect on model accuracy. Details are in Section~\ref{exp}. Even though DNN surrogate models perform well on average with randomly drawn input samples, they are susceptible to small changes in some directions. 

\subsection{Adversarial Training}
To enhance robustness against adversarial samples, a widely used approach is to add adversarial examples into the training data, known as {\it adversarial
training}~\citep{tramer2017ensemble}. The addition of adversarial samples is either literal~\citep{kurakin2016adversarial} or indirectly done by training with a modified loss function~\citep{goodfellow2014explaining}. We adopt the second approach because 
it is time consuming to generate the adversarial samples---we must query the original simulator with perturbed $x^{\prime}$ to obtain the output $y$. In particular, we propose the following adversarial loss function:
\begin{equation}
\widetilde{J}(\theta, x, y)=\alpha J(\theta, x, y)+(1-\alpha) J\left(\theta, x+\delta(x), y\right),
\label{loss}
\end{equation}
where $\delta(x)$ is the perturbation generated by the white-box adversarial attack methods FGSM or FGNM, $\alpha$ is the weight of the original loss (we set it to $0.8$ in our experiments). The second term in the adversarial loss function imposes smoothness on the DNN.
The rationale for (\ref{loss}) is that with a small amount of perturbation, the new output $F(x+\delta(x))$ should not move far away from the original output $y$, providing protection for the model under adversarial attacks. We explicitly show the dependence of $\delta(x)$ on $x$ to emphasize that it is not a single direction. By setting $\delta(x)$ in the most sensitive direction of the network at $x$, we ensure that the network is regularized most effectively. In other words, the use of $\delta(x)$ in (\ref{loss}) does not imply we seek robustness only along one direction given by $\delta(x)$. Also note that since $\delta(x)$ depends on the gradient of the trained DNN, it is updated iteratively as part of the optimization. Consequently, $\delta(x)$ computed from the final DNN based on the loss (\ref{loss}) is different from $\delta(x)$ based on the DNN without adversarial training. To evaluate the robustness of the final DNN, we will use the most sensitive perturbation directions of this DNN in addition to the most sensitive directions of the DNN without adversarial training. 

\section{Experiments}
\label{exp}
In this section, we use hypothesis testing to verify that the DNN surrogate models trained with the original loss are susceptible to changes in certain directions. Moreover, we demonstrate that the DNN surrogate models trained with the modified loss can yield higher accuracy on both the original test data and the adversarially perturbed data.

\subsection{Problem Setting}
We consider the benchmark elliptic partial differential equations (PDEs) on the $2$-d unit square domain in Eq. (\ref{eq:pde}), which is called Mindlin–Reissner (RM) plate model~\citep{mindlin1951influence,reissner1945effect}. We follow the parameter setting in \cite{luo2019deep}. Let $\Omega$ be a smooth domain, $\boldsymbol{C}$ a positive definite tensor, and $\varepsilon$ a linear green strain tensor. Use $\boldsymbol{\theta}=\left[\theta_{x}, \theta_{y}\right]$ to record the rotations of the surface and $\omega$ to specify the transverse displacement in z-direction. Parameter $\gamma$ is the scaled shear stresses; $f$ is the applied scaled transversal load; and $\lambda=\frac{E \kappa}{2(1+v)}$ is the shear modulus, where $E$ is Young's modulus, $v=0.3$ is the Poisson ratio, and $\kappa=\frac{5}{6}$ is the shear correction factor. In short, Eq. (\ref{eq:pde}) is a system of second order PDEs that describe a clamped plate bent by a transverse force.
\begin{equation}
\begin{aligned}
-\operatorname{div} \boldsymbol{C \varepsilon}(\boldsymbol{\theta})-\gamma=0 \quad &\text { in } \quad \Omega\\
-\operatorname{div} \gamma=f \quad &\text { in }\quad \Omega\\
-\lambda t^{-2}(\nabla \omega-\theta)=\gamma \quad &\text { in }\quad \Omega\\
\boldsymbol{\theta}=0, \omega=0 \quad &\text { on }\quad \partial \Omega
\end{aligned}
\label{eq:pde}
\end{equation}

Here the unknown input is the material field, i.e. Young's modulus $E \in \mathbb{R}^{64 \times 64}$, which is modeled as a log normal random field:
\begin{eqnarray*}
\log E(s) \sim G P\left(m(s), k\left(s, s^{\prime}\right)\right),
\end{eqnarray*}
where $m(s)$ and $k\left(s, s^{\prime}\right)$ are the mean and covariance functions of the Gaussian process. Based on the settings of~\cite{luo2019deep}, the mean function is zero and the  exponential kernel is adopted as the covariance function:
\begin{eqnarray*}
k\left(s, s^{\prime}\right)=\sigma^{2} \exp \left(-\frac{\left(s-s^{\prime}\right)^{2}}{2 l^{2}}\right) \, ,
\end{eqnarray*}
with the correlation length $l = 0.5$. In this paper, we study two datasets generated from the Mindlin–Reissner (RM) plate model, where the inputs are both Young's modulus $E$ and the outputs are two different fields: displacement field $D\in \mathbb{R}^{64 \times 64}$ and stress field $S\in \mathbb{R}^{64 \times 64 \times 3}$. The displacement field dataset is denoted by $\mathbb{D}$, and the stress field dataset is $\mathbb{S}$. The PDE simulator is solved by the finite element method (FEM)~\citep{hughes2012finite}.

For each dataset, we generate $2024$ samples in total and put aside $1000$ samples for testing. Following~\cite{luo2019deep}, for $\mathbb{D}$, we use a surrogate DNN, specifically, a convolutional neural network (CNN) containing $21$ layers, while for $\mathbb{S}$, we use a $25$-layer CNN. The prediction task for both datasets is regression, so the mean squared error (MSE) is used as the loss $J$. $L_2$ penalty, Mini-batch~\citep{cotter2011better}, and Adam optimizer~\citep{kingma2014adam} are used for training.

\subsection{Performance in Regression}
We now evaluate the regression performance of DNN surrogate models trained respectively with the original loss and the adversarial loss on the above two datasets. Denote the DNN surrogate model trained with the original loss (including $L_2$ penalty) $J(\theta, x, y)+\lambda \|\theta\|^2_2$ by $\text{DNN}_\text{ori}$, while the one with the adversarial loss $\widetilde{J}(\theta, x, y)+\lambda \|\theta\|^2_2$ by $\text{DNN}_\text{adv}$. We remind that the perturbation used in $\widetilde{J}$ is generated by FGNM (Eq.~\ref{fgnm}) with $\epsilon=0.1$. On each dataset, the two models are evaluated on $3$ types of test data: 
\begin{enumerate}
    \item 
    Original test data denoted by $\mathbb{D}_\text{test}$ and $\mathbb{S}_\text{test}$.
    \item
    Test data perturbed using the adversarial directions of $\text{DNN}_\text{ori}$, denoted by $\mathbb{DP}_\text{ori}$ and $\mathbb{SP}_\text{ori}$. 
    \item
    Test data perturbed using the adversarial directions of $\text{DNN}_\text{adv}$, denoted by $\mathbb{DP}_\text{adv}$ and $\mathbb{SP}_\text{adv}$. 
\end{enumerate}
The adversarial directions of any trained DNN are computed by FGNM and FGSM (Eq.~\ref{fgsm}), and the amount of perturbation is specified by $\epsilon$. Although we only present results at $\epsilon=0.1$, we also experimented with different values of $\epsilon$, {\it e.g.}, $0.01$ and $1$, and obtained similar results. For any perturbed input point, we compute the corresponding ``ground-truth'' output using the PDE simulator (Eq.~\ref{eq:pde}). Table~\ref{result} shows the regression results in terms of {\it mean squared error} (MSE). 
We can see that the accuracy of $\text{DNN}_\text{ori}$ is highly sensitive to the perturbation direction. When we use the adversarial direction computed by FGNM, for both datasets, compared with the not perturbed data, the MSE of $\text{DNN}_\text{ori}$ increases by nearly one order of magnitude or more. When FGSM is used to generate the adversarial directions, the increase of MSE is not as dramatic as with FGNM, but still more than $5$ times of that on the original data. However, if the perturbation is not targeted at the adversarial directions of $\text{DNN}_\text{ori}$, specifically, if the adversarial directions of $\text{DNN}_\text{adv}$ are used instead, the accuracy of $\text{DNN}_\text{ori}$ is close to that obtained on the original data. In contrast, the accuracy of $\text{DNN}_\text{adv}$ is much more stable across the original data, data perturbed in its adversarial directions, or the adversarial directions of $\text{DNN}_\text{ori}$. For both datasets, the worst MSE of $\text{DNN}_\text{adv}$ occurs with perturbation in the adversarial direction computed by FGSM, but the increase in MSE compared with that of the original data is below $18\%$. It is also interesting that the MSE values achieved by $\text{DNN}_\text{adv}$ for both original datasets are lower than those by $\text{DNN}_\text{ori}$, indicating that with much-enhanced robustness, the average accuracy of $\text{DNN}_\text{adv}$ is not compromised but instead improved.

\begin{table}
  \caption{Regression performance in MSE achieved by $\text{DNN}_\text{ori}$ and $\text{DNN}_\text{adv}$ on datasets $\mathbb{D}$ and $\mathbb{S}$.}
  \label{result}
  \centering
  \begin{tabular}{lllllll}
  \hline
   \multirow{2}{*}{} & \multirow{2}{*}{$\mathbb{D}_\text{test}$}  &  FGNM-$\mathbb{DP}_\text{ori}$ & FGNM-$\mathbb{DP}_\text{adv}$   \\
    &  &  (FGSM-$\mathbb{DP}_\text{ori}$) & (FGSM-$\mathbb{DP}_\text{adv}$)   \\
  \hline
   \multirow{2}{*}{$\text{DNN}_\text{ori}$} & \multirow{2}{*}{112.125} & 1806.347 & 104.831\\
    & & (1012.510) &(112.322) \\
   \multirow{2}{*}{$\text{DNN}_\text{adv}$} & \multirow{2}{*}{110.107} & 101.083 & 115.275 \\
    &  & (101.404) & (127.438) \\
  \hline
   \multirow{2}{*}{} & \multirow{2}{*}{$\mathbb{S}_\text{test}$}  &  FGNM-$\mathbb{SP}_\text{ori}$ & FGNM-$\mathbb{SP}_\text{adv}$   \\
    &  &  (FGSM-$\mathbb{SP}_\text{ori}$) & (FGSM-$\mathbb{SP}_\text{adv}$)   \\
  \hline
   \multirow{2}{*}{$\text{DNN}_\text{ori}$} & \multirow{2}{*}{855.567} & 7812.958 & 844.360\\
    &  & (4372.825) &(899.756) \\
   \multirow{2}{*}{$\text{DNN}_\text{adv}$} & \multirow{2}{*}{728.513} & 835.686 & 803.132 \\
    & & (825.421) &(855.130) \\
\hline
  \end{tabular}
\end{table}

\begin{figure*}[htbp!]
\centering
\begin{tabular}{ccc}
\includegraphics[width=2.0in]{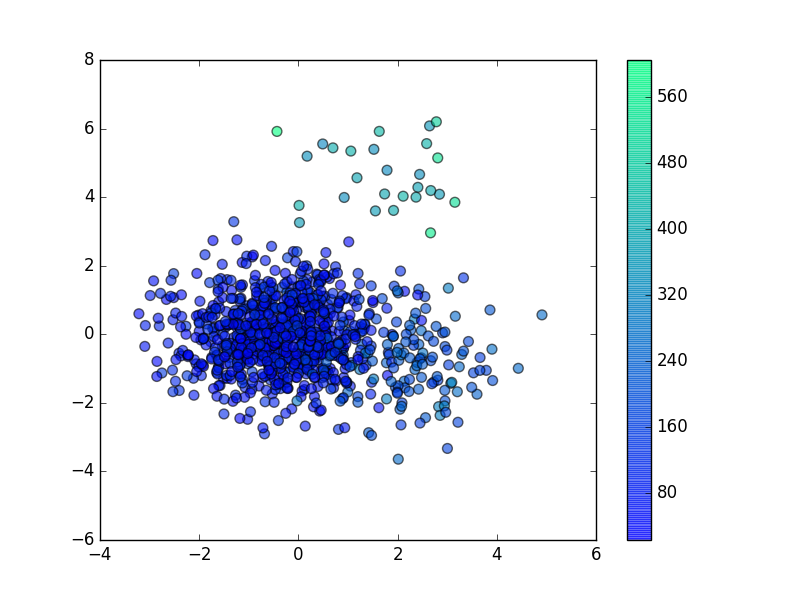} &
\includegraphics[width=2.0in]{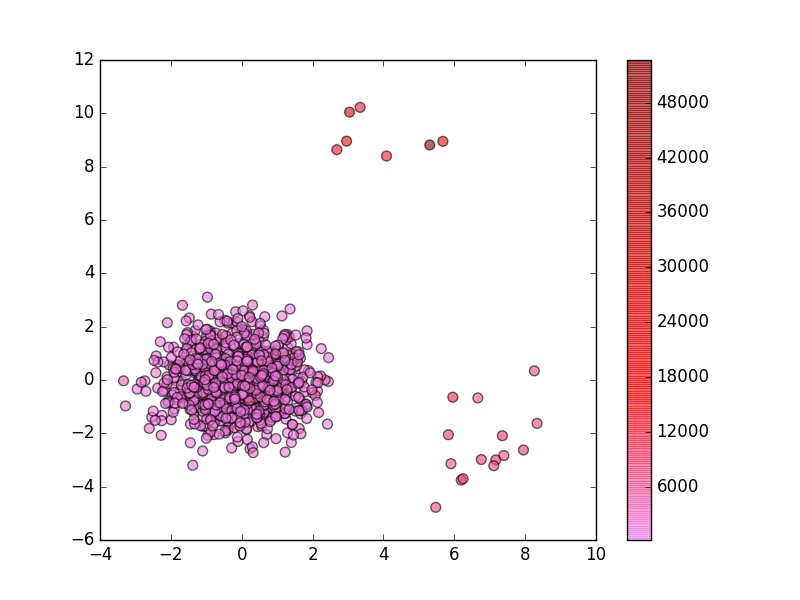} &
\includegraphics[width=2.0in]{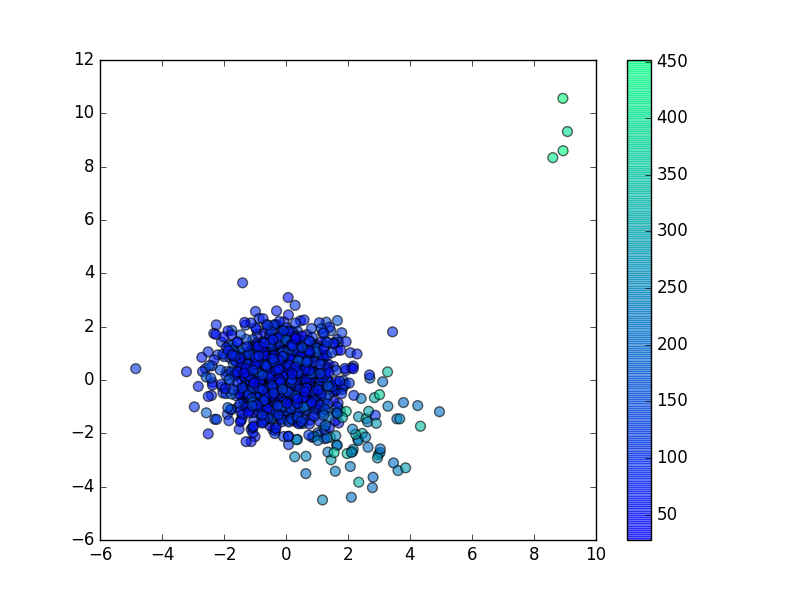}  \\
(a) & (b) & (c)\\
\includegraphics[width=2.0in]{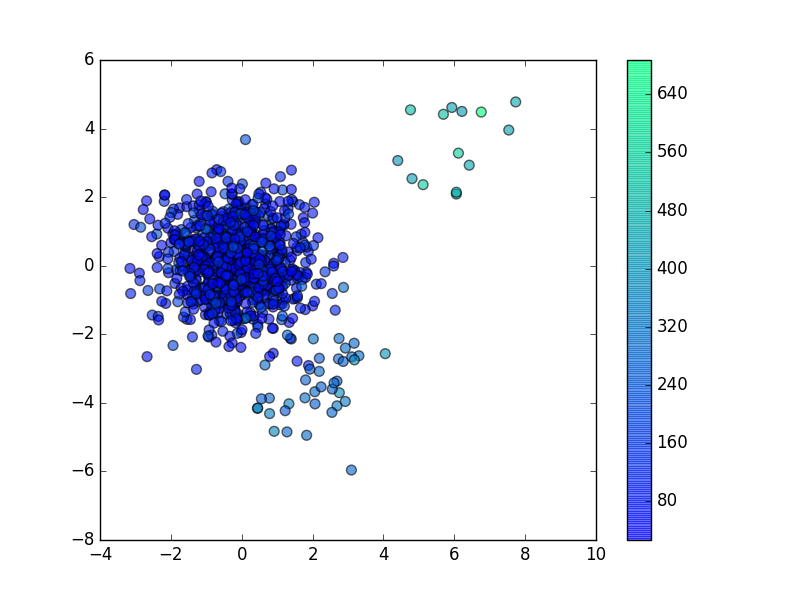} &
\includegraphics[width=2.0in]{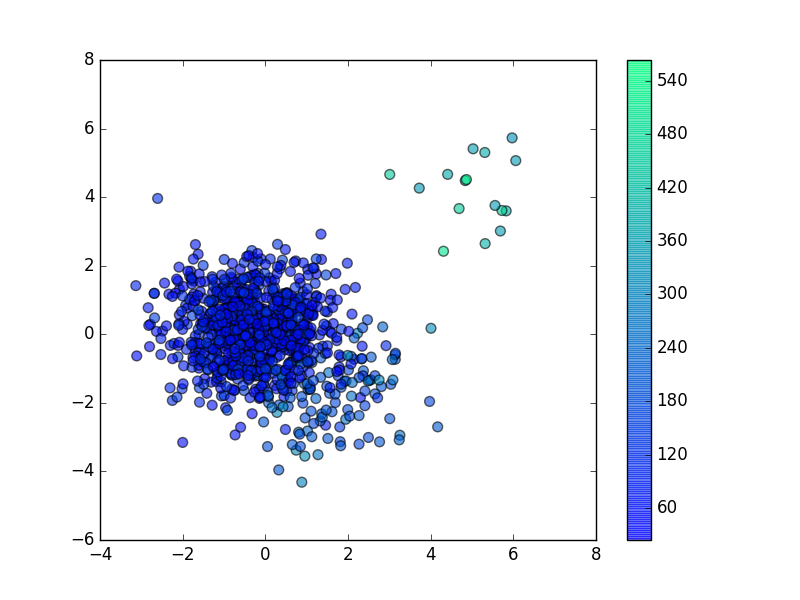} &
\includegraphics[width=2.0in]{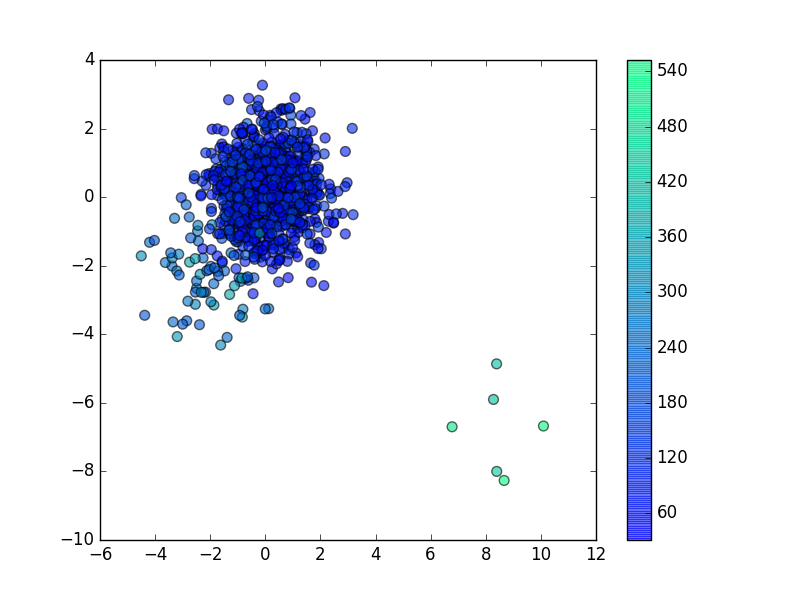}  \\
(d) & (e) & (f)\\
\includegraphics[width=2.0in]{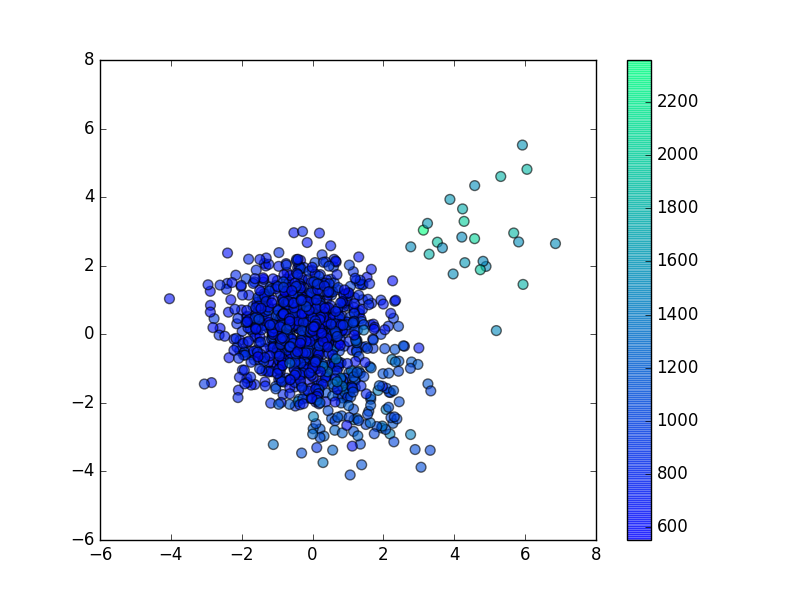} &
\includegraphics[width=2.0in]{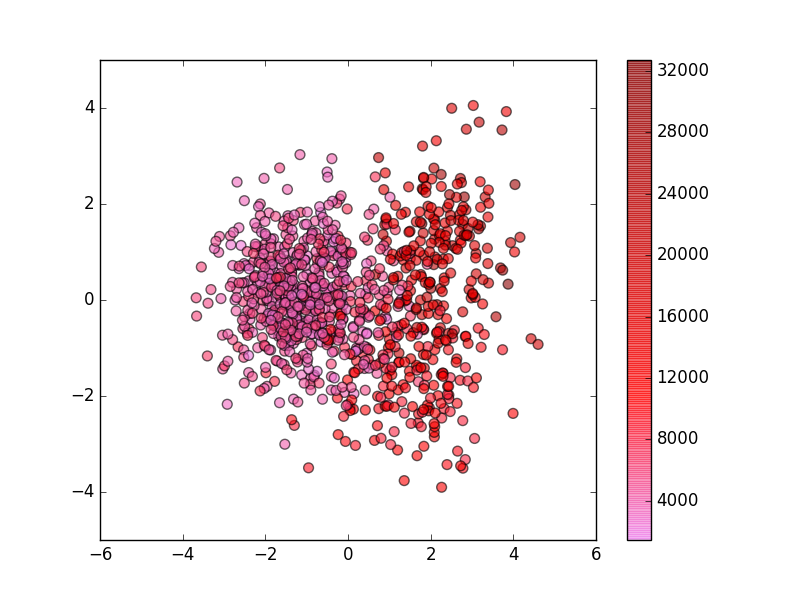} &
\includegraphics[width=2.0in]{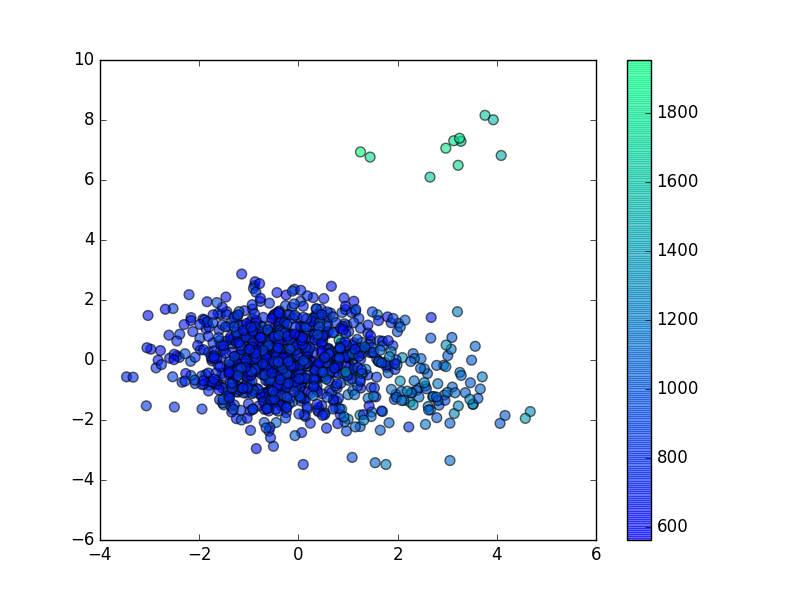}  \\
(g) & (h) & (i)\\
\includegraphics[width=2.0in]{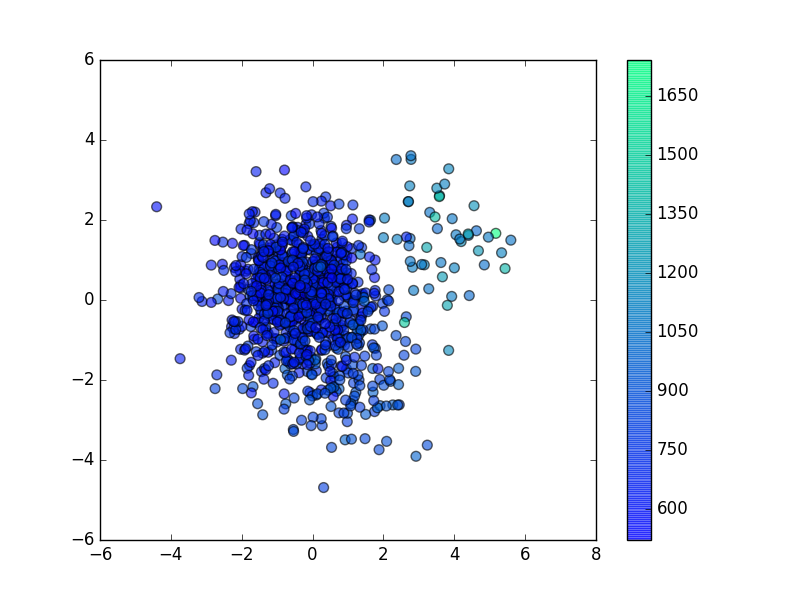} &
\includegraphics[width=2.0in]{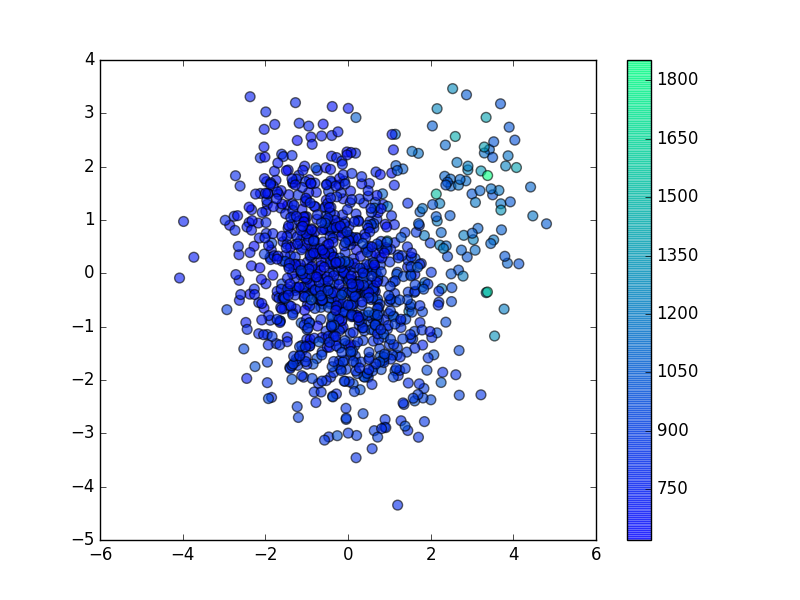} &
\includegraphics[width=2.0in]{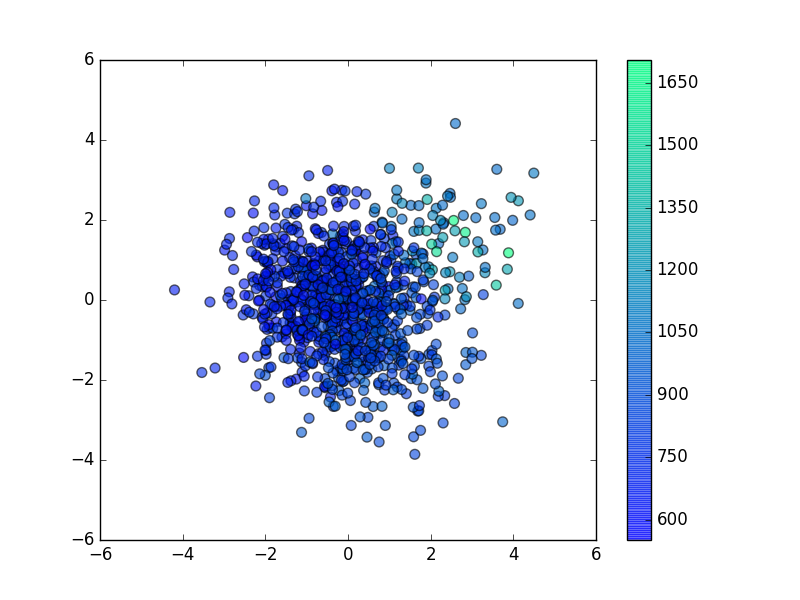}  \\
(j) & (k) & (l)
\end{tabular}
\caption{SE heatmap plots obtained by $\text{DNN}_\text{ori}$ and $\text{DNN}_\text{adv}$ on dataset $\mathbb{D}$ and $\mathbb{S}$. The horizontal and vertical coordinates are determined by LDA-based dimension reduction. Plots (a)-(f) are for $\mathbb{D}$ and (g)-(l) are for $\mathbb{S}$. (a) and (g): $\text{DNN}_\text{ori}$ on the original test data. (b) and (h): $\text{DNN}_\text{ori}$ on test data perturbed by FGNM based on $\text{DNN}_\text{ori}$. (c) and (i): $\text{DNN}_\text{ori}$ on test data perturbed by FGNM based on $\text{DNN}_\text{adv}$. (d) and (j): $\text{DNN}_\text{adv}$ on the original test data. (e) and (k): $\text{DNN}_\text{adv}$ on test data perturbed by FGNM based on $\text{DNN}_\text{ori}$. (f) and (l): $\text{DNN}_\text{adv}$ on test data perturbed by FGNM based on $\text{DNN}_\text{adv}$.}
\label{d}
\end{figure*}

The {\it squared errors} (SE) on the original test data and test data perturbed by FGNM are visualized in the heatmap plots in Figure~\ref{d}. The horizontal and vertical axes of the plots correspond to the two dominant coordinates computed by linear discriminant analysis (LDA) dimension reduction~\citep{balakrishnama1998linear}. To apply LDA, we divide the samples into three classes according to their SE values. The SE values are indicated by color. We see that for both datasets, compared with the performance on the original data, $\text{DNN}_\text{ori}$ 
yields much higher SE on the data perturbed in the adversarial direction calculated from the network itself.
For the data perturbed in the adversarial direction calculated from  $\text{DNN}_\text{adv}$, $\text{DNN}_\text{ori}$ performs similarly as with the original data. In contrast, $\text{DNN}_\text{adv}$ achieves consistent accuracy on all the test data, regardless of whether the data are perturbed or in which direction they are perturbed. 

\subsection{Effects of Adversarial Directions}
To demonstrate that the accuracy of $\text{DNN}_\text{ori}$ is highly sensitive to perturbation directions, but standard MC sampling can hardly detect such sensitivity, we conduct the following experiment.  
We randomly draw $50$ sample points from the $1000$ test data. For each point, we generate $100$ randomly perturbed points by adding Gaussian noise with mean zero and the same norm as the FGNM perturbation. The Gaussian noise ensures that the perturbation direction is uniformly random across all possible directions.
For every perturbed data point, we compute SE achieved by $\text{DNN}_\text{ori}$ and then fit a 1-D density plot of $\log(\mbox{SE})$ using the results for the $5000$ points. Specifically, the kernel density estimation (KDE)~\citep{davis2011remarks,parzen1962estimation}) is used. We obtain results at three levels of $\epsilon$ for datasets $\mathbb{D}$ and $\mathbb{S}$ respectively.
The perturbed versions of the dataset $\mathbb{D}$ at $\epsilon=\epsilon_i$ ($0.01, 0.1, 1$) are denoted by $\mathbb{D}_{\epsilon=\epsilon_i}$, and similar notations are used for those of $\mathbb{S}$. Denote by $f_{rand}(\cdot\mid \mathbb{D}_{\epsilon=\epsilon_i})$ the fitted density based on randomly perturbed $\mathbb{D}_{\epsilon=\epsilon_i}$. Likewise, we have $f_{rand}(\cdot\mid \mathbb{S}_{\epsilon=\epsilon_i})$. For comparison, we also fit two density functions for $\log(\mbox{SE})$ based on the $50$ samples perturbed at the same $\epsilon$ in the adversarial directions given respectively by FGNM and FGSM. Denote these density functions by $f_{FGNM}(\cdot\mid \mathbb{D}_{\epsilon=\epsilon_i})$ and $f_{FGSM}(\cdot\mid \mathbb{D}_{\epsilon=\epsilon_i})$.

\begin{figure*}[htbp!]
\centering
\begin{tabular}{ccc}
\includegraphics[width=1.9in]{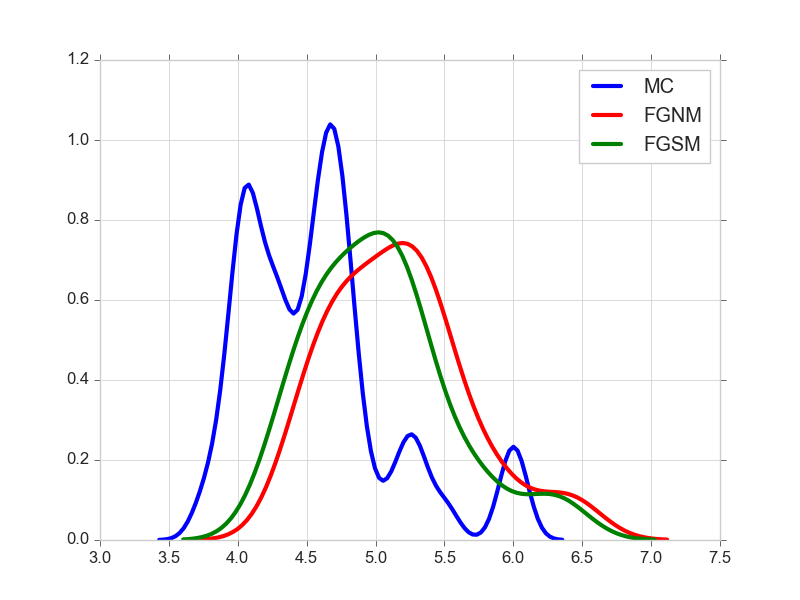} &
\includegraphics[width=1.9in]{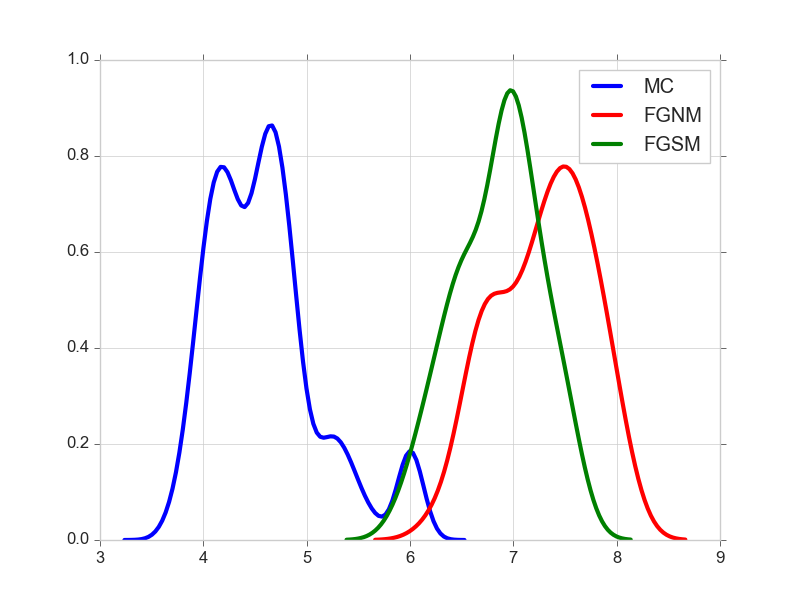} &
\includegraphics[width=1.9in]{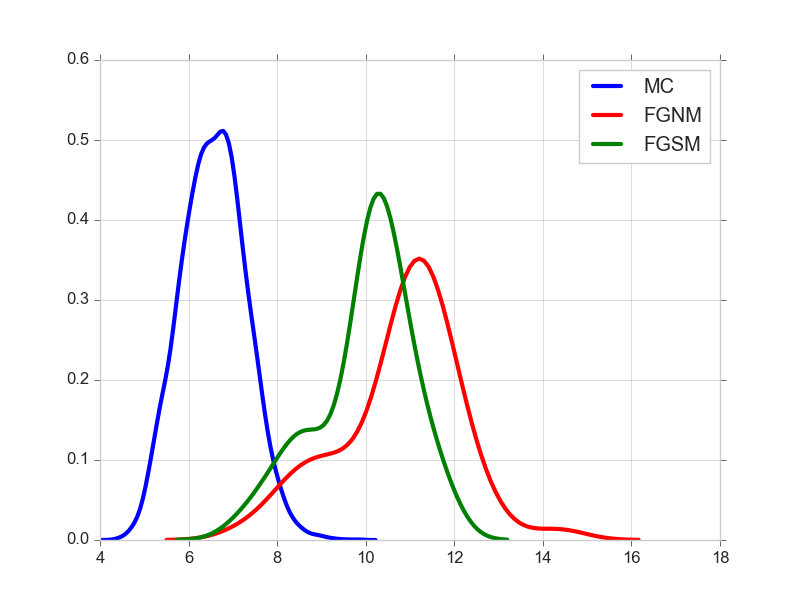}  \\
(a) & (b) & (c)\\
\includegraphics[width=1.9in]{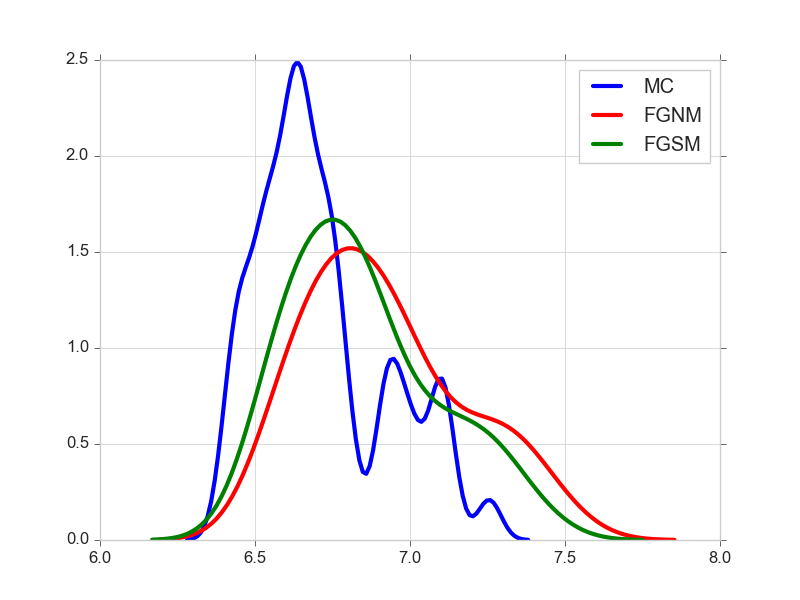} &
\includegraphics[width=1.9in]{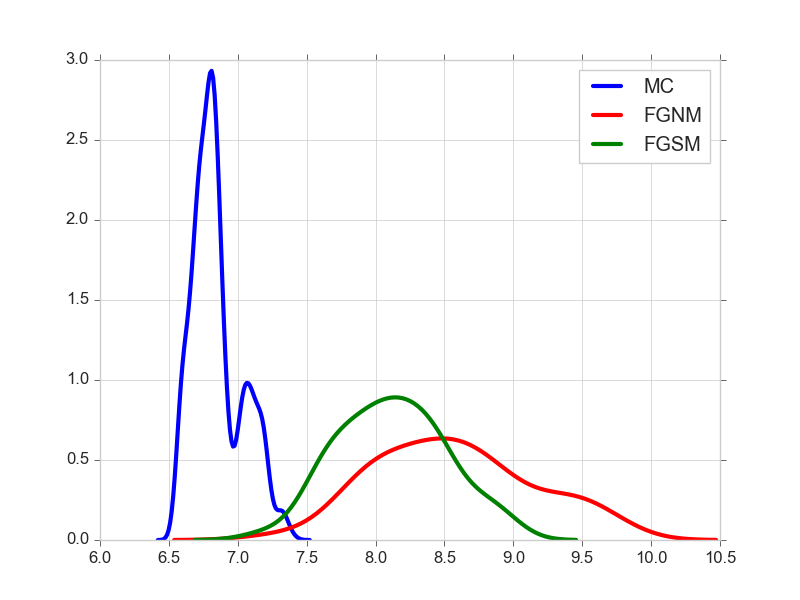} &
\includegraphics[width=1.9in]{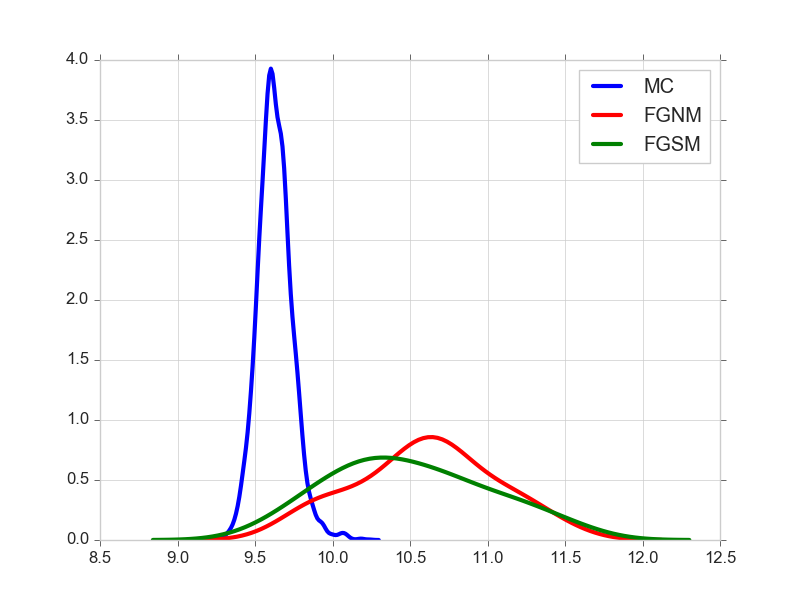}  \\
(d) & (e) & (f)
\end{tabular}
\caption{Density plots of $\log(\mbox{SE})$ obtained by $\text{DNN}_\text{ori}$ using MC samples generated by perturbing $50$ sample points. The value of $\epsilon$ determines the amount of perturbation. In each plot, the red and green lines correspond to the fitted density functions based on the $50$ random samples perturbed in the adversarial directions given by FGNM and FGSM respectively, while the blue line corresponds to the density based on the $50$ samples perturbed in $100$ randomly directions. (a)$\sim$(c): Dataset $\mathbb{D}$, $\epsilon=0.01, 0.1, 1$. (d)$\sim$(f): Dataset $\mathbb{S}$, $\epsilon=0.01, 0.1, 1$.}
\label{density}
\end{figure*}

Figure~\ref{density} shows the fitted density functions for every $\mathbb{D}_{\epsilon=\epsilon_i}$ and $\mathbb{S}_{\epsilon=\epsilon_i}$.
We see that $f_{FGNM}(\cdot\mid \mathbb{D}_{\epsilon=0.01} )$ and $f_{FGSM}(\cdot\mid \mathbb{D}_{\epsilon=0.01})$ overlap substantially with $f_{rand}(\cdot\mid \mathbb{D}_{\epsilon=0.01})$ although the average $\log(\mbox{SE})$ for either of the former two densities is higher than that of the latter (similar results for $\mathbb{S}_{\epsilon=0.01}$).
At larger values of $\epsilon$, i.e., $0.1$ or $1$, $f_{FGNM}(\cdot\mid \mathbb{D}_{\epsilon=\epsilon_i})$ and $f_{FGSM}(\cdot\mid \mathbb{D}_{\epsilon=\epsilon_i})$
locate mostly to the right of $f_{rand}(\cdot\mid \mathbb{D}_{\epsilon=\epsilon_i})$. The average of $\log(\mbox{SE})$ is significantly larger than the maximum value obtained on the samples perturbed in a random direction. This observation shows that evaluation based on the standard MC samples can grossly underestimate the error of the surrogate model in some directions. 
Techniques in adversarial learning are crucial for revealing this issue.

Next, we conduct hypothesis testing to investigate whether the performance of $\text{DNN}_\text{ori}$ is significantly worse in the adversarial directions compared with a random direction. We generate another test dataset by adding Gaussian noise (zero mean and the same norm as the FGNM perturbation) to the $1000$ test points. We calculate the MSE obtained by $\text{DNN}_\text{ori}$ on these randomly perturbed data and compare the result with that obtained from data perturbed in the adversarial directions. We apply FGNM instead of FGSM to determine the adversarial directions because the former has induced more increase in MSE according to Table~\ref{result}. Without making parametric assumptions about the distributions of MSE, we use the Mann–Whitney U test~\citep{mann1947test} (a non-parametric rank test) to test whether $\text{DNN}_\text{ori}$ performs equally well on the randomly perturbed data and the data perturbed in adversarial directions. The alternative hypothesis is that $\text{DNN}_\text{ori}$ yields a higher MSE on data perturbed in the adversarial directions. We perform the test on perturbed data generated at $\epsilon=0.01, 0.1, 1$ respectively. 
From Table~\ref{test}, we see that all the $6$ tests on $\text{DNN}_\text{ori}$ yield p-values lower than $0.001$. Thus at a significance level of $0.05$, the null hypothesis of every test is rejected. We conclude that $\text{DNN}_\text{ori}$ yields significantly worse MSE for data perturbed in the adversarial directions than data perturbed in a random direction.

We conduct the same hypothesis tests on $\text{DNN}_\text{adv}$ to see whether its accuracy changes significantly depending on the direction of perturbation. The adversarial directions are generated by FGSM instead of FGNM because the former results in a higher MSE for $\text{DNN}_\text{adv}$ according to Table~\ref{result}. Table~\ref{test} shows that for dataset $\mathbb{D}$, we cannot reject the null hypothesis at the significance level $\alpha=0.05$ when $\epsilon=0.01$. But there is a significant difference when $\epsilon$ is larger. For dataset $\mathbb{S}$, all the three tests have p-values greater than $0.05$, which indicates that MSE achieved by $\text{DNN}_\text{adv}$ has no significant difference between adversarial directions and randomly selected directions. Compared with $\text{DNN}_\text{ori}$, $\text{DNN}_\text{adv}$ is more robust to perturbation in adversarial directions. 

\begin{table}
  \caption{P-values of hypothesis testing on different magnitude of the perturbation.}
  \label{test}
  \centering
  \begin{tabular}{llll||lll}
  \hline
    & $\mathbb{D}_{\epsilon=0.01}$  &  $\mathbb{D}_{\epsilon=0.1}$ & $\mathbb{D}_{\epsilon=1}$  &
    $\mathbb{S}_{\epsilon=0.01}$  &  $\mathbb{S}_{\epsilon=0.1}$ & $\mathbb{S}_{\epsilon=1}$   \\
  \hline
   $\text{DNN}_\text{ori}$ & 0.000 & 0.000 &0.000 & 0.000 & 0.000 & 0.000 \\
   $\text{DNN}_\text{adv}$ & 0.152 & 0.000 &0.000 & 0.107 & 0.353 & 1.000 \\
\hline
  \end{tabular}
\end{table}

\subsection{Performance in Uncertainty Quantification}

\begin{table}
  \caption{Uncertainty Quantification by the first and second-order moment of every output variable. Relative errors (REs) are listed to measure the difference between a surrogate model and the simulator.}
  \label{uncertainty1}
  \centering
  \begin{tabular}{lllllll}
  \hline
   \multirow{3}{*}{} & $\mathbb{D}_\text{test}$  &  $\mathbb{DP}_\text{rand}$  & $\mathbb{DP}_\text{fg}$   \\
    & $1$st Moment &  $1$st Moment & $1$st Moment  \\
    & ($2$nd Moment) &  ($2$nd Moment) & ($2$nd Moment)   \\
  \hline
   \multirow{2}{*}{$\text{DNN}_\text{ori}$} & 0.001 & 0.001 & 0.004\\
    & (0.001) & (0.001) &(0.007) \\
   \multirow{2}{*}{$\text{DNN}_\text{adv}$} & 0.000 & 0.000 & 0.000 \\
    & (0.001) & (0.001) & (0.001) \\
  \hline
   \multirow{3}{*}{} & $\mathbb{S}_\text{test}$  &  $\mathbb{SP}_\text{rand}$  & $\mathbb{SP}_\text{fg}$   \\
    & $1$st Moment &  $1$st Moment & $1$st Moment   \\
    & ($2$nd Moment) &  ($2$nd Moment) & ($2$nd Moment)   \\
  \hline
   \multirow{2}{*}{$\text{DNN}_\text{ori}$} & 0.001 & 0.001 & 0.009 \\
    & (0.001) & (0.001) &(0.009) \\
   \multirow{2}{*}{$\text{DNN}_\text{adv}$} & 0.001 & 0.001 & 0.001 \\
    & (0.001) & (0.001) & (0.001) \\
\hline
  \end{tabular}
\end{table}

\begin{figure*}[htbp]
\centering%
\begin{tabular}{cccc}
\includegraphics[width=1.8in]{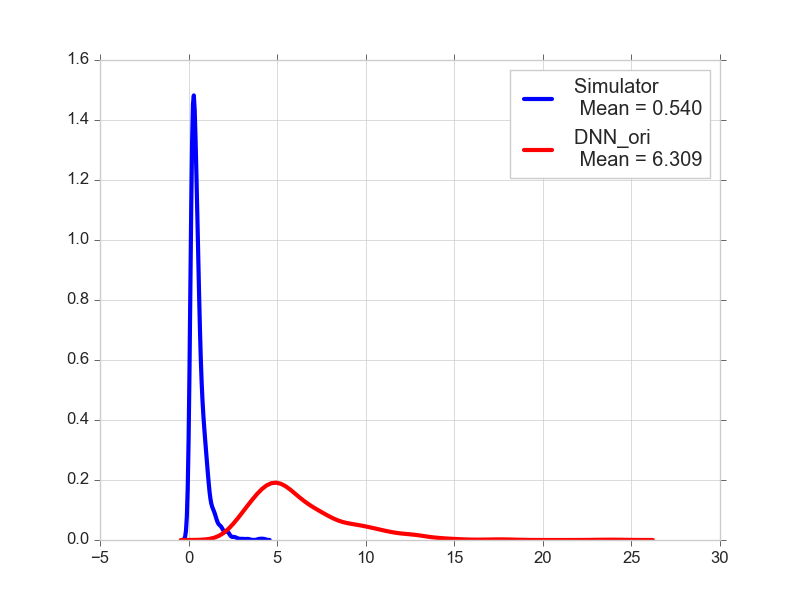} &
\includegraphics[width=1.8in]{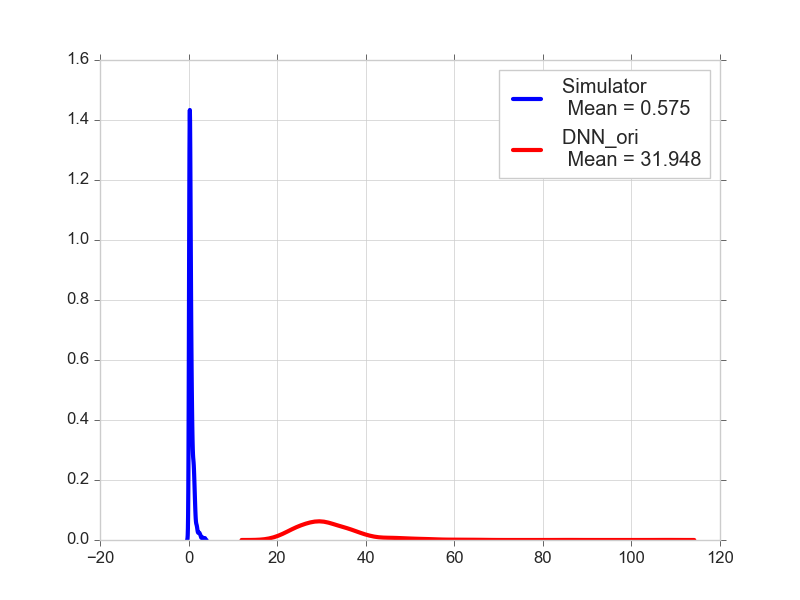} \\
(a) & (b) \\
\includegraphics[width=1.8in]{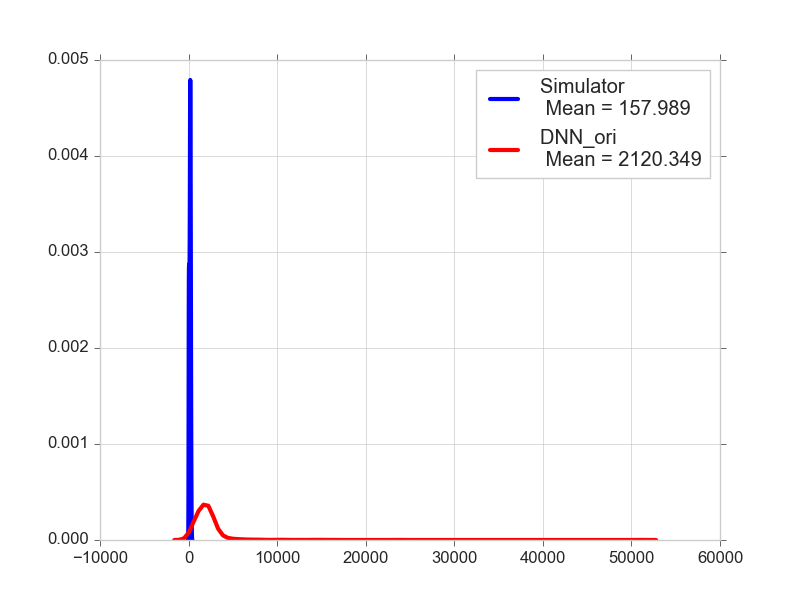} &
\includegraphics[width=1.8in]{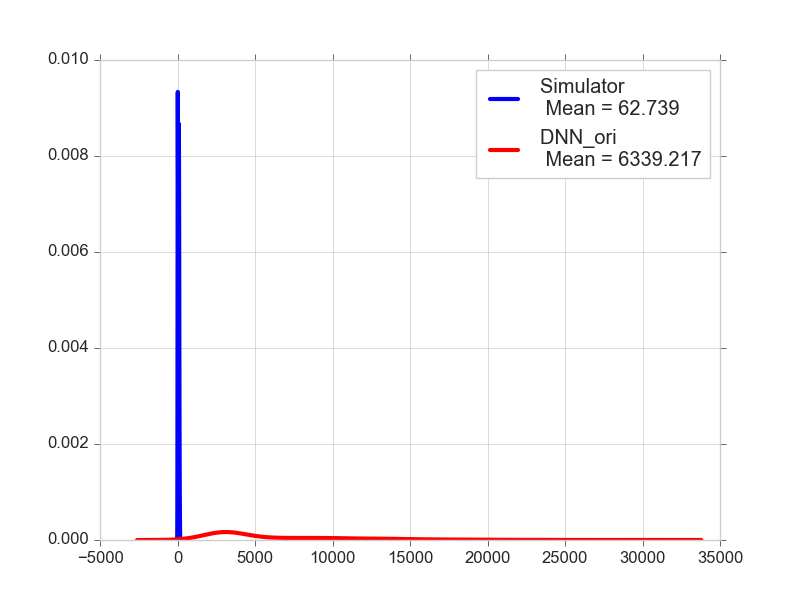} \\
(c) & (d) \\
\includegraphics[width=1.8in]{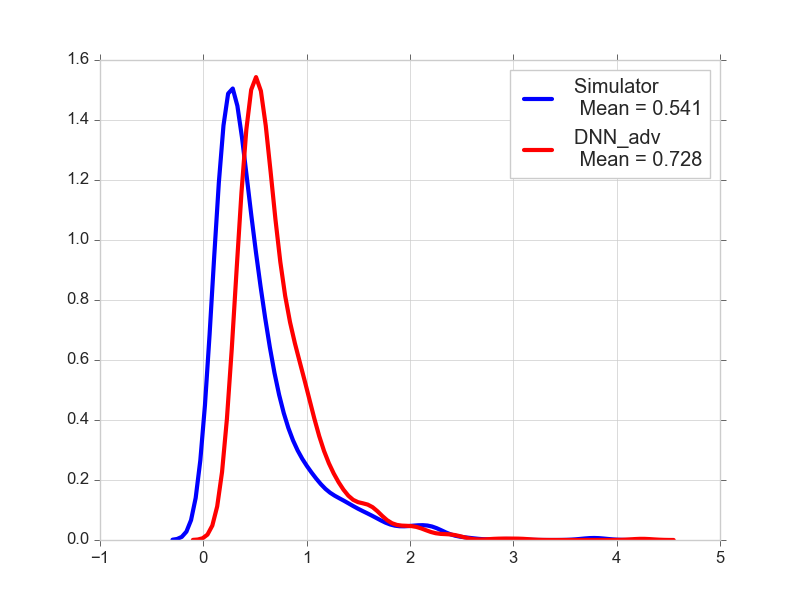} &
\includegraphics[width=1.8in]{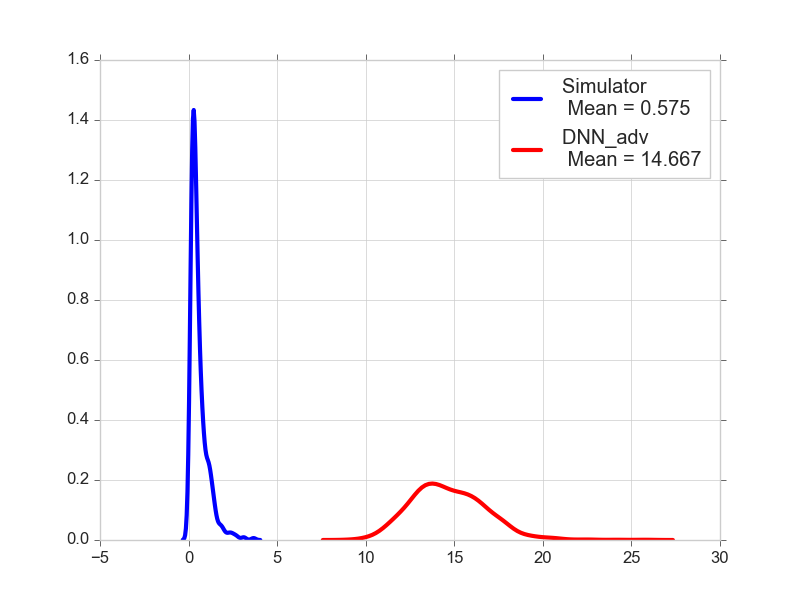}\\
(e) & (f)\\
\includegraphics[width=1.8in]{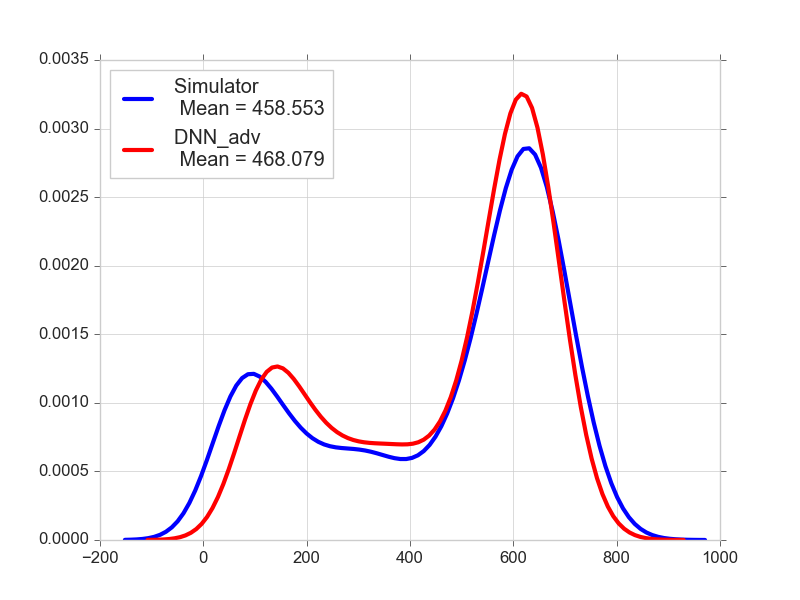}  &
\includegraphics[width=1.8in]{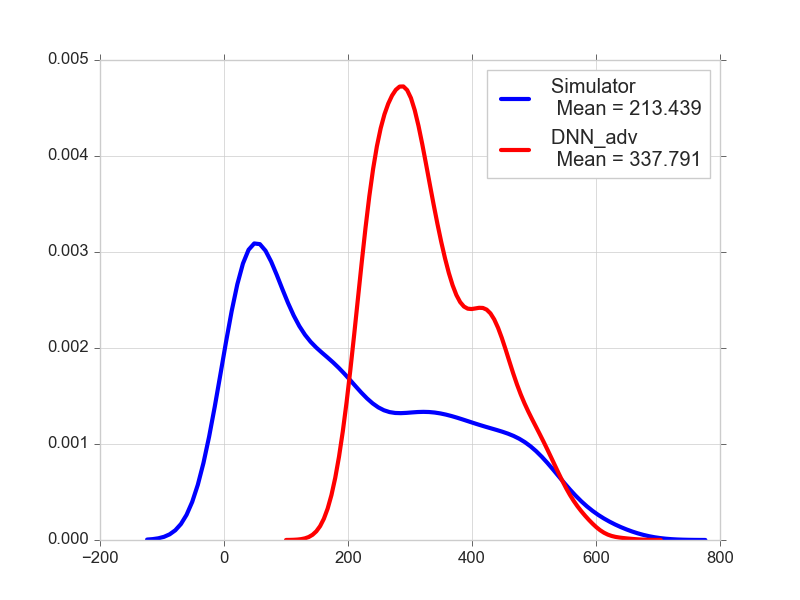}  \\
(g) & (h)\\
\end{tabular}
\caption{Density plots of $SE$ between the outputs of the original data and perturbed data obtained from two datasets $\mathbb{D}$ and $\mathbb{S}$. In each plot, the blue line shows the density of $SE$ when the simulator is used, while the red line shows the density when $\text{DNN}_\text{ori}$ or $\text{DNN}_\text{adv}$ is used. The mean SE of each distribution is given in the plots. These plots are obtained by different choices of datasets, ways of perturbing the input, and the surrogate model used. Plots in the left column (a, c, e, g) are for $\mathbb{D}$ and those in the right column (b, d, f, h) are for $\mathbb{S}$. (a) \& (b): Data perturbed in random direction, comparison between the simulator and $\text{DNN}_\text{ori}$. (c) \& (d): Data perturbed in the adversarial direction, comparison between the simulator and $\text{DNN}_\text{ori}$. (e) \& (f): Data perturbed in random direction, comparison between the simulator and $\text{DNN}_\text{adv}$. (g) \& (h): Data perturbed in the adversarial direction, comparison between the simulator and $\text{DNN}_\text{adv}$. 
}
\label{uncertainty2}
\end{figure*}

Depending on the application, we may be interested in different kinds of UQ. We consider two practices here. First, treating the input as random variables, we regard the output variables as random. Statistics of each output variable, often, its first and second-order moments are computed. These statistics inform us of the expected value and the expected amount of variation at the output when the input is sampled from a given distribution. We anticipate that an accurate surrogate model produces similar UQ statistics as the original simulator. We hereby calculate the first and second-order moments of each output variable using the PDE simulator, $\text{DNN}_\text{ori}$, and $\text{DNN}_\text{adv}$. For both $\mathbb{D}$ and $\mathbb{S}$, the two surrogate models are evaluated on $3$ types of test data ($\epsilon$ is set to $0.1$):
\begin{enumerate}
    \item 
    Original test data denoted by $\mathbb{D}_\text{test}$ and $\mathbb{S}_\text{test}$.
    \item
    Test data perturbed using random directions, denoted by $\mathbb{DP}_\text{rand}$ and $\mathbb{SP}_\text{rand}$. 
    \item
    Test data perturbed using the adversarial directions of the corresponding model, denoted by $\mathbb{DP}_\text{fg}$ and $\mathbb{SP}_\text{fg}$. Specifically, for $\text{DNN}_\text{ori}$, FGNM is used to decide the adversarial direction, and for $\text{DNN}_\text{adv}$, FGSM is used. For clarity of the presentation, we only report results obtained by FGNM perturbation for $\text{DNN}_\text{ori}$ because the experiments show that the drop in accuracy by FGSM perturbation is less severe. For $\text{DNN}_\text{adv}$, the opposite is true---FGSM perturbation causes more degradation in accuracy. Since we are examining the robustness of these networks, we present the results for the case of the stronger adversarial effect.
\end{enumerate}

As the output for the two datasets is either a 2-D or 3-D array, we obtain an array of the first or second-order moment for any model. Denote the array of a moment obtained by the simulator by $m_\text{sim}$ and that by a DNN surrogate model by $m_\text{DNN}$. 
The overall disparity in the first or second-order moment between the surrogate model and the simulator is measured by the {\it relative error} ({\it RE}): 
\[
\mathcal{E} = \| m_\text{DNN} - m_\text{sim} \|_F/\| m_\text{sim} \|_F \, ,
\]
where $\| \cdot \|_F$ denotes the Frobenius norm. Results are provided in Table~\ref{uncertainty1}. We see that $\text{DNN}_\text{adv}$ outperforms $\text{DNN}_\text{ori}$, as shown by smaller RE values across all datasets. Moreover, for data perturbed in the adversarial directions, the RE obtained by $\text{DNN}_\text{ori}$ increases sharply while that by $\text{DNN}_\text{adv}$ stays the same as with randomly perturbed data.



Secondly, we consider uncertainty caused by imprecise input. A natural question is how much the output would change when the input deviates from the truth to a certain extent.  To estimate the variation, outputs at perturbed inputs are computed by the simulator. The SE between the outputs with or without input perturbation is then calculated. We can also fit a density for the SE values based on many perturbed inputs. For the sake of UQ, we would like to have a surrogate model that generates similar distributions of SE.
For both datasets $\mathbb{D}$ and $\mathbb{S}$, we evaluate $\text{DNN}_\text{ori}$ and $\text{DNN}_\text{adv}$ by calculating SE values in two cases ($\epsilon$ is set to $0.1$): 
\begin{enumerate}
    \item 
    Between the outputs of any original input and a randomly perturbed input. 
    \item
    Between the outputs of any original input and its perturbation in the adversarial direction.
\end{enumerate}
Figure~\ref{uncertainty2} shows comparisons of the distributions of SE between each surrogate model and the simulator. We see that regardless of whether the direction of perturbation is random or adversarial, the distributions of SE differ significantly between $\text{DNN}_\text{ori}$ and the simulator, while $\text{DNN}_\text{adv}$ has yielded more similar density functions as the simulator. For $\text{DNN}_\text{ori}$ versus the simulator, even the support ranges of the distributions barely overlap. For $\mathbb{D}$, $\text{DNN}_\text{adv}$ yields distributions similar in both shape and the range of support to those by the simulator. In summary, for both tasks of UQ, $\text{DNN}_\text{adv}$ outperforms $\text{DNN}_\text{ori}$.

\section{Conclusions}
\label{conclude}
In this paper, we raise the awareness of a significant drawback of DNN surrogate models---the emulation accuracy can drop significantly when the input is perturbed slightly in a direction determined by the gradient of the network. This trait of DNNs has motivated active research on adversarial training but has not attracted due attention from researchers using surrogate models. We demonstrate the severity of this problem using hypothesis testing. We also show that this problem is not easily detectable based on standard Monte Carlo sampling. By exploiting techniques in adversarial training, we have developed an approach to improve the robustness of DNN surrogate models. Experiments show that the DNNs trained by the new method perform substantially better with adversarial samples in terms of both emulation accuracy and uncertainty quantification. Furthermore, the new method achieves slightly better average emulation accuracy.

\section*{Acknowledgments}

The research is supported by the National Science Foundation under grant DMS-2013905.

\bibliography{ref.bib}

\end{document}